# Performance Prediction and Optimization of Solar Water Heater via a Knowledge-Based Machine Learning Method

Hao Li
*University of Texas at Austin, United States*

Zhijian Liu
*North China Electric Power University, China*

**ABSTRACT**

*Measuring the performance of solar energy and heat transfer systems requires a lot of time, economic cost and manpower. Meanwhile, directly predicting their performance is challenging due to the complicated internal structures. Fortunately, a knowledge-based machine learning method can provide a promising prediction and optimization strategy for the performance of energy systems. In this Chapter, the authors will show how they utilize the machine learning models trained from a large experimental database to perform precise prediction and optimization on a solar water heater (SWH) system. A new energy system optimization strategy based on a high-throughput screening (HTS) process is proposed. This Chapter consists of: i) Comparative studies on varieties of machine learning models (artificial neural networks (ANNs), support vector machine (SVM) and extreme learning machine (ELM)) to predict the performances of SWHs; ii) Development of an ANN-based software to assist the quick prediction and iii) Introduction of a computational HTS method to design a high-performance SWH system.*

**TERMS AND DEFINITIONS**

SWH: solar water heater

ANN: artificial neural network

SVM: support vector machine

ELM: extreme learning machine

HTS: high-throughput screening

WGET-SWH: water-in-glass evacuated tube solar water heater

MLFN: multilayer feedforward neural network

HCR: heat collection rate

HLC: heat loss coefficient

TCD: tube center distance

RMS: root mean square

PC: personal computer

## INTRODUCTION

Predicting the thermal performance of solar energy systems is of huge challenge due to the complexity of the internal structures. In fact, the measurement of thermal performances of a typical solar energy system (*e.g*., solar water heater (SWH)) requires a lot of time, economic costs and labors. Due to the complicated structures, conventional physical and mathematical models usually fail to estimate their thermal performances. These problems not only dramatically hinder the acquisition of the thermal performances of solar energy systems, but also block the possibility of optimizing their thermal performance.

In the past decades, scientists have come up with a powerful prediction method to address these problems. People found that a knowledge-based machine learning model can help precisely predict the performances of some energy systems utilizing some simple independent variables as the computational inputs. With a proper machine learning algorithm, people only need to acquire a sufficient experimental database as well as perform the model training and testing, and then a predictive model can be acquired. During the training process, machine learning can "learn" the non-linear relationship between the independent and dependent variables via a "black box" fitting, and subsequently perform the predictions. The research group of Prof. S. Kalogirou, from Cyprus University of Technology, has conducted a majority of the pioneer application research on the prediction of thermal performances for energy systems (Kalogirou, 1999), leading to huge positive engineering impacts during the last two decades. Subsequently, relevant studies have become increasingly popular all over the world. Meanwhile, there are more and more new or revised machine learning algorithms developed. Among various machine learning approaches, there are some most widely used algorithms, such as artificial neural network (ANN) (Kalogirou, 1999), support vector machine (SVM) (Suykens & Vandewalle, 1999) and extreme learning machine (ELM) (Huang et al., 2006). ANN is the most prevalent algorithm due to its long history and powerful predictive capacity. The general schematic structure of an ANN model is presented in Figure 1.

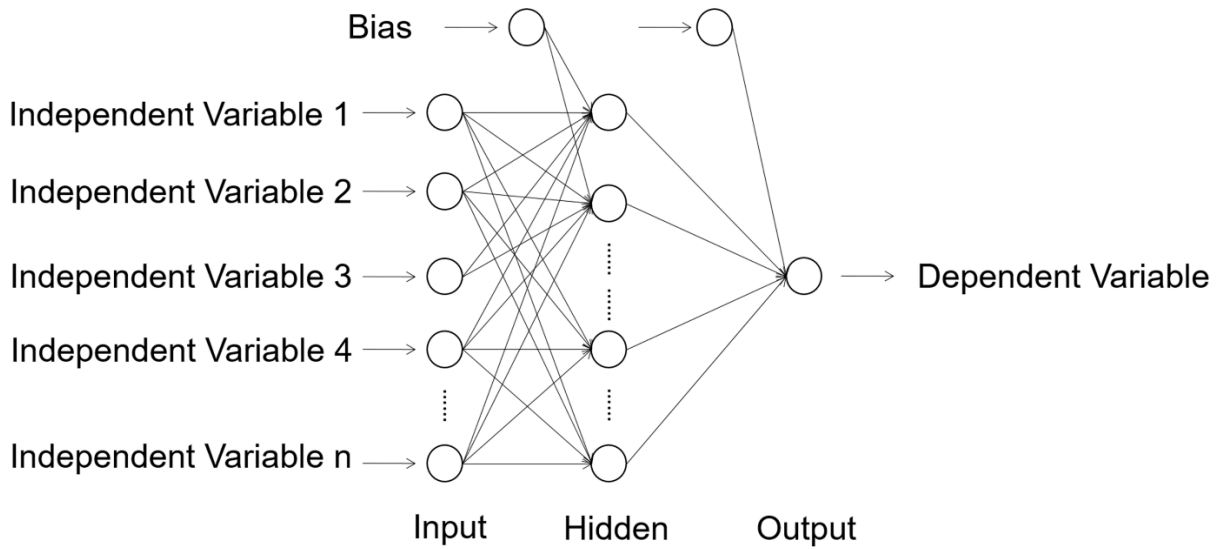

*Figure 1. General schematic structure of a typical ANN model. The empty circles represent the neurons. All the neurons interconnect with other neurons in the adjacent layer(s). Each neuron in the input layer represents an independent variable. The neuron in the output layer represents the dependent variable. Reproduced with permission from Reference* (Li, Liu, Liu, & Zhang, 2017).

So far, despite the great progress of machine learning, engineering and industrial requirements have been rising in recent years: how to cost-effectively design and optimize a solar energy system by utilizing machine learning? Now, machine learning is a proven powerful tool for varieties of numerical predictions, and people are trying to make full use of its predictive power, as well as provide a good optimization strategy in order to acquire higher performances. Nevertheless, to the best of the authors' knowledge, very few research reports have mainly focused on the relevant studies (Peng & Ling, 2008). Recently, the authors have found that with a sufficiently large experimental database, the machine learning models are not only able to give excellent predictive performance for SWHs, but also assist an efficient and promising optimization of the thermal performances of SWHs, with the usage of a high-throughput screening (HTS) process (Zhijian Liu, Li, Liu, Yu, & Cheng, 2017). This is, so far, the first study of HTS on the optimization of energy system.

SWH is one of the most popular techniques where the solar collectors and concentrators are employed to gather, store, and utilize solar radiation to heat air or water (Mekhilef, Saidur, & Safari, 2011). Among several different types of stationary collectors, the evacuated tube solar collectors are featured by the edges of lowering the heat loss coefficient and lower economic cost than other conventional flat plate collectors. In lots of countries, the all-glass evacuated tubular solar water heaters are particularly popular due to the good thermal performance, easy installation, and good transportability. The annual production of evacuated solar tubes kept growing in lots of developing countries (*e.g.*, China) and the market share of them kept increasing at rapid speed during the past decades (Tang, Li, Zhong, & Lan, 2006).

In this Chapter, all the studies will focus on a typical SWH, the water-in-glass evacuated tube solar water heater (WGET-SWH, Figure 2). The reasons why WGET-SWH was selected as a case study are as below: i) WGET-SWH is one of the most common SWHs in developing countries; ii) experimental measurements of WGET-SWH's thermal performances is time-consuming and tedious; and iii) several most significant thermal properties (*e.g.*, heat collection rate and heat loss coefficient) of WGET-SWH have not been well-studied. In the authors' previous studies, the properties of 915 WGET-SWHs were experimentally measured according to a National Standard of China (GB/T 19141)(Zhijian Liu, Li, Zhang, Jin, & Cheng, 2015). The measured properties include heat collection rates (daily heat collection per square meter of a solar water system, $MJ/m^2$), heat loss coefficients (the average heat loss per unit, $W/(m^3K)$), tube length (mm), number of tubes, tube center distance (mm), tank volume (maximum mass of water in tank, kg), collector area ($m^2$), tilting angle (°) and final temperature (°C). Except the heat collection rate and heat loss coefficient, all other properties were measured by the "portable test instruments" (Table 1). Descriptive statistics of the measured data (maximum, minimum, range, average and standard deviation) are shown in Table 2.

According to the standard of measurements, measuring the thermal properties of a WGET-SWH requires around 15 days and a series of tedious setups. To provide a quick alternative for the measurement, the authors have developed a machine learning-based method to directly predict the heat collection rate and heat loss coefficient of a WGET-SWH with the inputs which can be measured from the "portable test instruments" (Zhijian Liu, Li, et al., 2015). In other words, once the machine learning model is properly trained with the use of the database, it will be able to predict quickly and precisely the heat collection rate and heat loss coefficient of a WGET-SWH. Such a novel measurement will help effectively reduce the measurement time from weeks to seconds, and thus will dramatically accelerate the measurements of SWHs in both industrial and commercial applications. More details will be introduced in the following Sections.

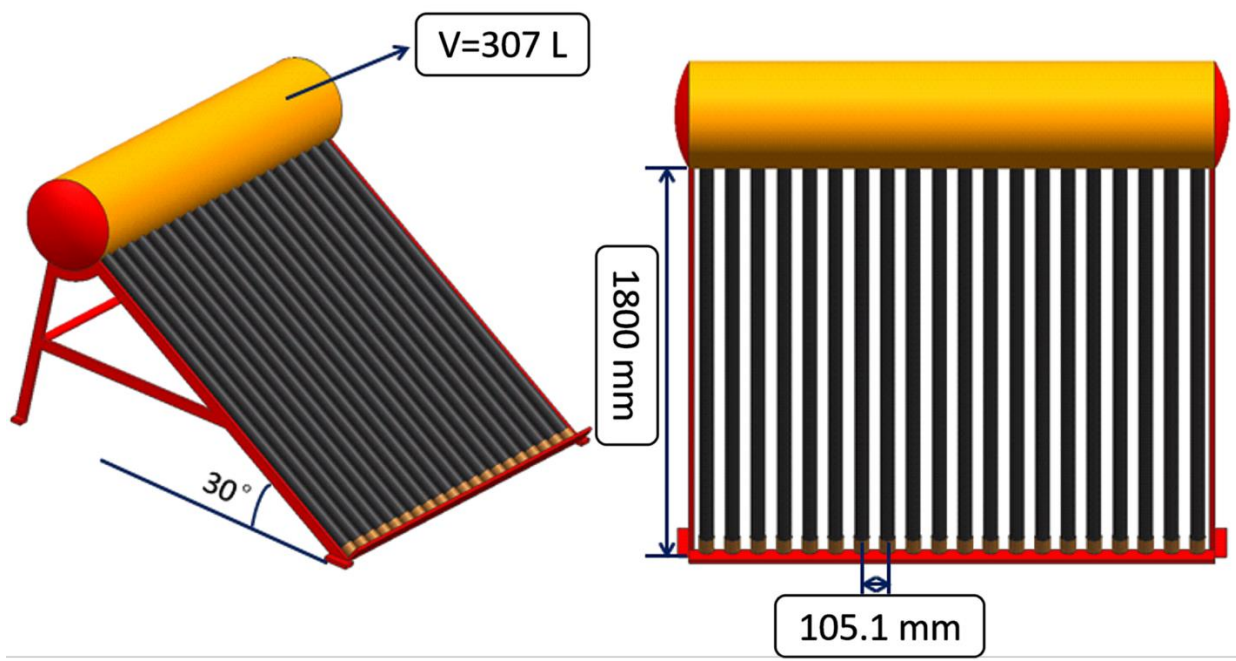

*Figure 2. Schematic picture of a representative WGET-SWH. Reproduced with permission from Reference* (Zhijian Liu, Li, Liu, et al., 2017).

*Table 1. "Portable test instruments" for measuring the properties of WGET-SWHs. Reproduced with permission from Reference* (Zhijian Liu, Li, et al., 2015).

| Parameters | Portable Test instruments | Accuracy |
|---|---|---|
| Final temperate of water | Digital thermoelectric thermometer | ±0.5% |
| Tank Volume | Electric platform scale | ±1.0% |
| Diameter, tube center distance, tube length, collector area | Taper ZSH-3 | ±0.5% |

*Table 2. Descriptive statistics of the variables for 915 samples of WGET-SWHs. Reproduced with permission from Reference* (Zhijian Liu, Li, et al., 2015).

| Items | Tube Length (mm) | Number of Tubes | TCD (mm) | Tank Volume (kg) | Collector Area ($m^2$) | Angle (°) | Final Temp. (°C) | HCR ($MJ/m^2$) | HLC ($W/(m^3K)$) |
|---|---|---|---|---|---|---|---|---|---|
| Maximum | 2200 | 64 | 151 | 403 | 8.24 | 85 | 62 | 11.3 | 13 |
| Minimum | 1600 | 5 | 60 | 70 | 1.27 | 30 | 46 | 6.7 | 8 |
| Range | 600 | 59 | 91 | 333 | 6.97 | 55 | 16 | 4.6 | 5 |
| Average | 1811 | 21 | 76.2 | 172 | 2.69 | 46 | 53 | 8.9 | 10 |
| Standard deviation | 87.8 | 5.8 | 5.11 | 47.0 | 0.73 | 3.89 | 2.0 | 0.48 | 0.77 |

Abbreviations: TCD: tube center distance, final temp.: final temperature, HCR: heat collection rate, $MJ/m^2$, HLC: heat loss coefficient ($W/(m^3K)$). Tank volume was defined as the maximum mass of water in tank (kg).

As a part of the handbook, here, the authors aim to present the predictive power of machine learning in the prediction and optimization of the thermal performances of SWHs, as well as pick the research of WGET-SWH as a case study. The content of this Chapter consists of three parts: i) Comparative studies on varieties of machine learning models (ANNs, SVM and ELM) to predict the thermal performances of SWHs; ii) Development of an ANN-based software to assist the quick and accurate prediction and iii) Introduction of a computational HTS strategy to design and optimize a high-performance SWH system.

## BACKGROUND

### Machine Learning Models

Machine learning is a powerful technique for numerical prediction, classification and pattern recognition, which has been widely used in chemical (Li et al., 2016; Li, Chen, Cheng, Zhao, & Yang, 2015), medical (Wernick, Yang, Brankov, Yourganov, & Strother, 2010), biological (Sommer & Gerlich, 2013), environmental (Zhijian Liu, Li, & Cao, 2017) and energy areas (Jordan & Mitchell, 2015; Sun, He, & Chang, 2015). In the authors' recent studies, various machine learning models were used to predict the heat collection rate and heat loss coefficient of WGET-SWH, such as ANN (Zhijian Liu, Liu, et al., 2015; Zhijian Liu, Li, et al., 2015), SVM (Zhijian Liu, Li, et al., 2015) and ELM (Z. Liu et al., 2016). In terms of ANN, several typical network algorithms were used, including general regression neural network (GRNN) (Specht, 1991) and multilayer feedforward neural network (MLFN) (Hornik, Stinchcombe, &

White, 1989). Because this handbook is for readers with broad interests, details of their algorithms are not discussed in this Chapter. General principles of ANN, SVM and ELM can be referred to References (Huang, Zhou, Ding, & Zhang, 2012; Kalogirou, 1999; Suykens & Vandewalle, 1999).

Before developing a predictive machine learning model, the inputs and output(s) of the model should be rationally selected. The inputs of the models should be the independent variables that are related (or partially related) to the dependent variable(s), while the output(s) should be the dependent variable(s) that should be predicted. Since one of the basic missions of machine learning is to help people acquire the knowledge that are hard to be measured or observed, it is strongly recommended that the dependent variable(s) should be the properties that are experimentally hard to be detected (if the expected dependent variable(s) can be easily acquired from experiments or can be precisely predicted by multiple linear regression or any physical models, bring in machine learning again would be a waste of time).

Development of a machine learning model for numerical prediction consists of the training and testing of datasets. The training process is essentially the process of a non-linear fitting (also called a "black-box" fitting). A good training of the dataset means that the training is neither under- nor over-fitting. Usually, if the dataset for training is not sufficiently large, there will be a huge risk of over-fitting. To ensure a good training result, a high percentage of training set is usually necessary. The testing process is a process to validate if the trained model is good, utilizing the dataset that is not previously involved in the training process. By comparing the data in the testing set (also called the "actual values") with the data predicted by the training set (also called the "predicted values"), people can calculate the root mean square errors (RMS errors), prediction accuracies (with a given tolerance) and residual values. The RMS error, prediction accuracy and residual value can be calculated by Equations (1), (2) and (3), respectively:

$$\boldsymbol{RMS\ error} = \sqrt{\frac{\Sigma_{i=1}^{n}(Z_i - O_i)^2}{n_{tot}}} \qquad (1)$$

$$\boldsymbol{Prediction\ accuracy} = \frac{n_{good}}{n_{tot}} \times \boldsymbol{100}\% \qquad (2)$$

$$\boldsymbol{Residual\ value} = Z_i - O_i \qquad (3)$$

where $Z_i$ and $O_i$ are the predicted and actual values, respectively; $n_{tot}$ is the number of tested samples; $n_{good}$ is the number of tested samples with good predicted results under a given tolerance. Empirically, the tolerance is usually set as ±30%.

A well-trained model is usually accompanied by the testing result with low RMS error, high prediction accuracy and low absolute values of residual. To ensure a good target model, comparing the performances of models with different training and testing percentages is necessary. On one hand, if the percentage of training set is too high, the results given by the testing set would not be reliable. On the other hand, if the percentage of training set is too low, there is a risk of over-fitting. To further show the availability of the model, a cross-validation and/or sensitivity test should be performed (Browne, 2000). However, cross-validation usually requires extremely high computational cost. If the database is very large, a regular personal computer (PC) will no longer be effective. Fortunately, it is found that if both the training and testing datasets are sufficiently large, with a robust and stable ANN algorithm (*e.g.*, GRNN), a cross-validation process can be rationally skipped after a simplified sensitivity test.

To sum up, Table 3 shows a suggested model development process. Good training and testing processes guarantee that the machine learning model can be used for practical applications. For practical applications, people only need to acquire and input the independent variables into the model, and then the dependent variable(s) will be predicted and outputted automatically.

*Table 3. A recommended machine learning development process for numerical prediction.*

|        | Steps                                                            | Notes                                                                                              |
| ------ | ---------------------------------------------------------------- | -------------------------------------------------------------------------------------------------- |
| Step 1 | Preparation of experimental database                             | The database should be sufficiently large.                                                         |
| Step 2 | Selection of independent and dependent variables                 | Independent variables should be easily-accessible.                                                 |
| Step 3 | Model training using the training set                            | Different percentages of training and testing sets should be tried.                                |
| Step 4 | Model testing using the testing set                              |                                                                                                    |
| Step 5 | Calculation of RMS error, prediction accuracy and residual values | If the testing results are not acceptable, the training should be performed again with different settings. |
| Step 6 | Cross-validation and/or sensitivity test (if necessary)          | NA                                                                                                 |
| Step 7 | Practical applications of the well-trained model                 | NA                                                                                                 |

## High-Throughput Screening (HTS)

HTS is generally defined as a method for experimentation previously used in medical and biological sciences (Hertzberg & Pope, 2000). With some state-of-the-art devices, algorithms and/or machines, an HTS process can help us quickly screen thousands or even millions of candidates (*e.g.*, chemical compounds (Hautier, Fischer, Jain, Mueller, & Ceder, 2010), materials (Greeley, Jaramillo, Bonde, Chorkendorff, & Nørskov, 2006), genes (Colbert, 2001) and biological designs (An & Tolliday, 2010; Wahler & Reymond, 2001)) with specific target performances for practical or scientific use. At first, the HTS was just used for drug discovery. But now it has been widely used in various areas, such as computer-assisted design of materials. The rough concept of a computational HTS process is pretty simple: the calculations of all possible candidates in a short timescale (using fast algorithms) and the screening of candidates with acceptable target performances (Li et al., 2017). Previously, Greeley and his colleagues used density functional theory (DFT) calculations to screen and design thousands of bimetallic catalysts for chemical reactions via an HTS process (Greeley et al., 2006). The predicted performances of their target design were in excellent agreement with experimental validations. Ceder and his colleagues combined DFT calculations, machine learning and an HTS method to screen all the possible ternary oxide compounds in the periodic table (Hautier et al., 2010). A large database of the nature's missing ternary oxide compounds was then developed. This study indicated that a machine learning-assisted HTS process can potentially be a good method to discover human's unknown knowledge. However, though these studies successfully used the concept of HTS for new knowledge searching, in the following years, there are very few relevant studies in the engineering area, especially the field of energy system design and optimization.

In the previous Sections of this Chapter, it is shown that machine learning can potentially be a good method for thermal performance optimization of SWH. Meanwhile, it is known that the design and optimization of a high-performance SWH is particularly difficult due to its complicated internal structure. Thus, we can come up with a "bold" idea: can we use a machine learning-based HTS process to predict

"infinite" possible designs of SWHs and screen the good candidates for practical use? The answer to this question will be given in the following Sections.

## THERMAL PERFORMANCE PREDICTION AND OPTIMIZATION

### Prediction of Thermal Performance

To find out a good machine learning model for predicting the thermal performances of WGET-SWHs, the authors have developed a series of models with the algorithms of GRNN, MLFN, SVM and ELM, respectively. The average RMSE errors and average prediction accuracies calculated from the testing results (after multiple training and testing processes) are shown in Table 4. The prediction accuracies were calculated with the tolerance of ±30%. It should be noted that all the data used for model training and testing are in the same database, as shown in Table 2. In terms of the MLFN and ELM, only the results of the best network configurations (optimal numbers of hidden layer and hidden neuron) are shown here. Surprisingly, with the tolerance of ±30%, all the four models have shown extremely good predictive performances for both heat collection rate and heat loss coefficient, with all the average prediction accuracies reach 100% (Table 4).

*Table 4. Predictive performances and modeling information of GRNN, MLFN, SVM and ELM for the predictions of the heat collection rate and heat loss coefficient for WGET-SWHs. Data source: References from* (Z. Liu et al., 2016; Zhijian Liu, Li, et al., 2015).

| Model | Average RMS Errors (for heat collection rate) | Average RMS Errors (for heat loss coefficient) | Percentages of Training and Testing Sets | Average Prediction Accuracy for Heat Collection Rate | Average Prediction Accuracy for Heat Loss Coefficient |
|---|---|---|---|---|---|
| GRNN | 0.33 | 0.71 | Training set: 85%; Testing set: 15%; | 100% | 100% |
| MLFN | 0.14 | 0.73 | Training set: 85%; Testing set: 15%; | 100% | 100% |
| SVM | 0.29 | 0.73 | Training set: 85%; Testing set: 15%; | 100% | 100% |
| ELM | 0.30 | 0.67 | Training set: 85%; Testing set: 15%; | 100% | 100% |

To show the training and testing results of the machine learning model, here, the authors pick one of the typical modeling results of an MLFN model, for the prediction of heat collection rate of WGET-SWHs (Figures 3 and 4, respectively). It can be clearly seen that, with a good ANN training process (Figure 3), the model can precisely predict the heat collection rates of the 137 data samples in a testing set (Figure 4), with relatively low absolute residual values. It should be noted that in the testing results, there are always some data points, which, more or less, deviate from the diagonal (Figure 4a). This is certainly acceptable—as gold cannot be pure and men cannot be perfect!

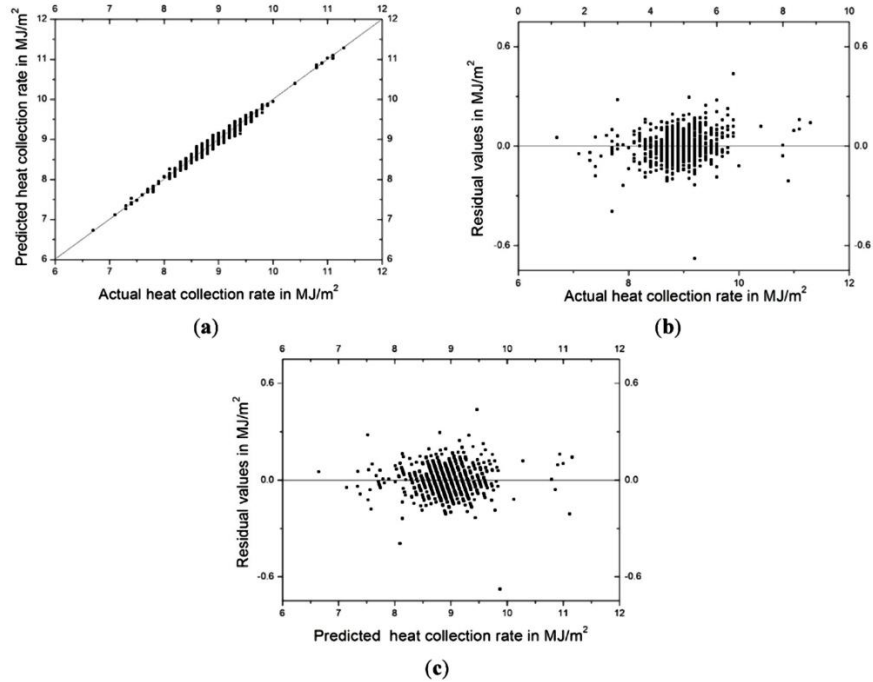

*Figure 3. Training results of 778 samples using an MLFN model for the prediction of heat collection rate for WGET-SWHs. (a) Predicted values vs actual values; (b) residual values vs actual values; (c) residual values vs predicted values. Reproduced with permission from Reference* (Zhijian Liu, Li, et al., 2015).

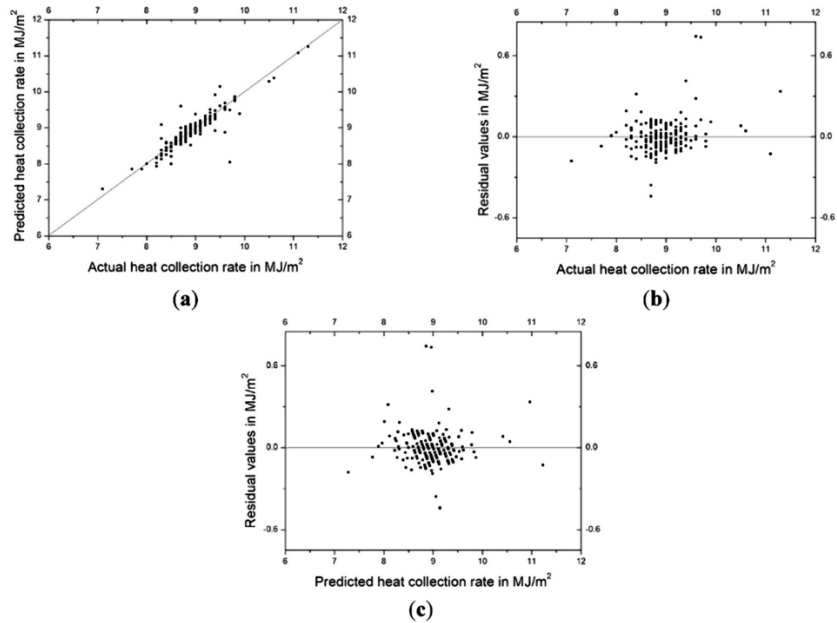

*Figure 4. Testing results of 137 samples using an MLFN model for the prediction of heat collection rate for WGET-SWHs. (a) Predicted values vs actual values; (b) residual values vs actual values; (c) residual values vs predicted values. Reproduced with permission from Reference* (Zhijian Liu, Li, et al., 2015).

Although the performances of all the four models look quite similar (as shown in Table 4), there are still several criterions that can help us distinguish which model is the most practical one. First is the training time. As we may know, compared to GRNN and SVM, MLFN and ELM have the structures that are closer to a conventional ANN, containing undefined numbers of hidden layer and/or hidden neurons. This means that people have to compare different network configurations (*e.g.*, different number of hidden layer and/or hidden neurons) in order to get the best algorithmic structure for prediction. Compared to GRNN and SVM (which only require the training once and for all), MLFN and ELM are clearly not the best options for this case study. In terms of the comparison between GRNN and SVM, both of them have the advantage of quick training and precise prediction. It is relatively hard to decide which one is the best since they have rather similar predictive performance in this study.

So far, the conclusion here is clear: machine learning is an effective method to predict the thermal performances of WGET-SWHs. Though it is very powerful, it is still far away from real applications: officers/workers in a company or industry usually don't want to spend a lot of money to learn and purchase a machine learning or Mathlab software— either a user-friendly machine learning package or Mathlab is rather expensive. So here is a new question: how to really use this method for practical application? In the next Section, a recently-developed ANN-based software will be introduced.

## Developing an ANN-Based Software

To provide an effective support for practical applications, the authors developed an user-friendly package, *WaterHeater*, in both PC and Android platforms (Zhijian Liu, Liu, et al., 2015). The primary motivation of this study was to provide a software that could help people quickly acquire the ANN-predicted heat collection rate and heat loss coefficient of WGET-SWHs, with the simple inputs measured by the "portable test instruments" (Table 1). The reason why the authors also developed an Android-based version was that a mobile system is sometimes more user-friendly and applicable for industrial measurement (people do not always have their computers aside). Though in different platforms, there is no difference between the models inside the packages of the PC and the Android versions. With this software, people only need to use a computer or even a mobile phone to perform quick thermal measurements utilizing the "portable test instruments". The inner core of the package is a well-trained MLFN with a back-propagation algorithm and a Sigma function as the activation function. The inputs are the same as the machine learning models described in previous sections. The predictive performance of the packages was validated by the residual values calculated from the testing set (Figure 5). It should be noted that though this model has some slight deviations for predicting the heat loss coefficients when their actual values are very high or very low (Figure 5b), the general predictive performance of the network work is still acceptable for the prediction of heat loss coefficient (because most of the heat loss coefficient in the industry are around 10 W/(m$^3$K), which is neither too high nor too low). In this Section, the application of this software will be introduced. The details of software developments can be found in Reference (Zhijian Liu, Liu, et al., 2015).

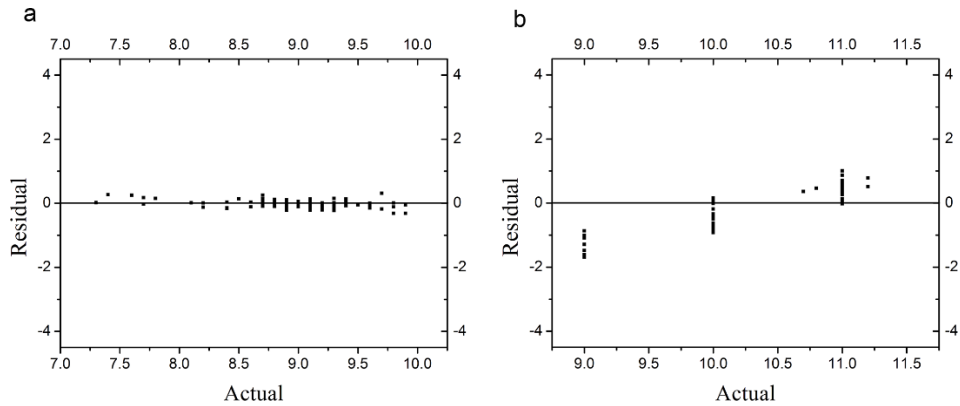

*Figure 5. Residual values vs. actual values in predicting (a) heat collection rates (in MJ/m$^2$) and (b) heat loss coefficient (in W/(m$^3$K)). All the actual values came from a testing set. Reproduced with permission from Reference* (Zhijian Liu, Liu, et al., 2015).

Figures 6 and 7 show the main panels of the software in PC and Android platforms, respectively. Both these two platforms consist of three parts: i) input panels, ii) output panels and iii) buttons. As their name imply, the input panel is the place for inputting independent variables, while the output panel is the place to display the dependent variables. The buttons consist of the "Reset Parameters" button and the "Start to Predict" button. By clicking the "Start to Predict" button with the values inputted in the input panel, the predicted heat collection rate and heat loss coefficient will be instantly shown in the output panel. By simply clicking the "Reset Parameters" button, all the input and output data will be erased.

*Figure 6. Overview panel of "WaterHeater" in a PC platform. Reproduced with permission from Reference* (Zhijian Liu, Liu, et al., 2015).

*Figure 7. Overview panel of "WaterHeater" in an Android platform. Reproduced with permission from Reference* (Zhijian Liu, Liu, et al., 2015).

This software can provide a perfect solution for quick thermal performance estimation of WGET-SWHs. Here, a quick measurement flow chart is proposed, as shown in Figure 8. With simple measurements of independent variables using the "portable test instruments" (Table 1), fast prediction of the thermal performance of a WGET-SWH can be achieved by the software. By measuring the tube length, tube center distance, tank volume, final temperature and titling angle (between tubes and ground), the heat collection rate and heat loss efficient can be quickly predicted when all these variables are inputted into the software panel. It should be mentioned that this ANN-based method does not aim to replace the conventional measurements. Instead, it provides a quick choice to predict the thermal values for the industrial and commercial uses that require rapid estimations of thermal performances. It should be noted that, to more precisely assess the thermal performance of a WGET-SWH, standard measurements should be used (though sometimes they are very time-consuming). This software is available at http://t.cn/RLPKF08 and for the latest version, reader can contact the corresponding author of this Chapter .

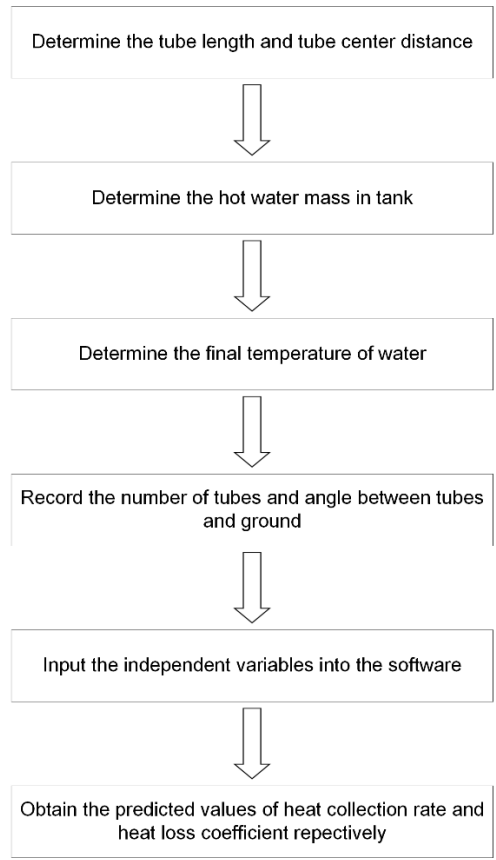

*Figure 8. Flow chart of the novel method using the "portable test instruments" combined with the software "WaterHeater" for the predictions of heat collection rate and heat loss coefficient. Reproduced with permission from Reference* (Zhijian Liu, Liu, et al., 2015).

## Optimizing the Thermal Performance via an HTS Strategy

It has been shown that machine learning can be such a powerful tool to predict the thermal performance of WGET-SWHs, here comes a new question: can people use this technique to predict the thermal performance of a newly designed WGET-SWH without direct experiments? The answer is yes. And it can be done in an even crazier way: screening thousands or millions of design candidates by using a well-trained machine learning model, and then selecting the candidates with good target performances. This screening strategy, as mentioned in the previous sections, is called the HTS process.

Using the optimization of heat collection rates of WGET-SWHs as a case study, the authors recently found that an HTS process with a proper ANN can be used for this mission. Usually, a high-performance WGET-SWH should have the heat collection rate as high as possible. For the screening process, the first step was to generate a large number of independent variable combinations (around $3.5 \times 10^8$ possible design combinations) as the inputs of the previously-trained ANN. Since the final temperature is not a part of the SWH installation, all the integers of final temperature between 52 and 62 °C were selected as the input. With all these independent variables (except the final temperature), people can easily construct a completed WGET-SWH in industry. The heat collection rates of the WGET-SWHs with all these possible combinations were then numerically predicted and outputted. The designed WGET-SWHs with high predicted heat collection rates will be screened and collected. For validation, the authors also experimentally installed two selected candidates and measured their thermal performances. Here, the two selected designs are respectively called "Design A" and "Design B", with the design and predicted details shown in Table 6. The predicted heat collection rates of these two designs are relatively high, no matter with which final temperature between 52 and 62 °C. As a result, both Designs A and B showed high average heat collection rates after standard measurements, as shown in Table 7. It should be noted that the environmental conditions (*e.g.*, solar radiation intensity, ambient temperature, season and location) for measuring these two designs are very similar to all the WGET-SWHs in the authors' database. That is to say, these two designs are comparable with the previous 915 WGET-SWHs. Surprisingly, it was found that the two designs had the average heat collection rate higher than all the 915 WGET-SWHs in the previous database. In the following content, details about this screening process will be introduced.

*Table 6. Predicted variables of two designed WGET-SWHs. Reproduced with permission from Reference* (Zhijian Liu, Li, Liu, et al., 2017).

|  | Tube Length (mm) | Number of Tubes | TCD (mm) | Tank Volume (kg) | Collector Area ($m^2$) | Angle (°) | Final Temp. (°C) |
|---|---|---|---|---|---|---|---|
| Design A | 1800 | 18 | 105.5 | 163 | 1.27 | 30 | 52-62 |
| Design B | 1800 | 20 | 105.5 | 307 | 1.27 | 30 | 52-62 |

Abbreviations: TCD: tube center distance, final temp.: final temperature. Tank volume was defined as the maximum mass of water in tank (kg).

*Table 7. Measured heat collection rates ($MJ/m^2$) of the two new designs. Reproduced with permission from Reference* (Zhijian Liu, Li, Liu, et al., 2017).

|  | Day 1 | Day 2 | Day 3 | Day 4 | Average | Predicted | Error rate |
|---|---|---|---|---|---|---|---|
| Design A | 11.38 | 11.26 | 11.34 | 11.29 | 11.32 | 11.47 | 1.35% |
| Design B | 11.47 | 11.43 | 11.42 | 11.45 | 11.44 | 11.66 | 1.90% |

Being similar to a previous computational HTS concept proposed by Aspuru-Guzik and his colleagues (Pyzer-Knapp, Suh, Gómez-Bombarelli, Aguilera-Iparraguirre, & Aspuru-Guzik, 2015), a modified HTS process for the optimization in this case has been proposed, as shown in Figure 9. Details about modeling and experimental contents can be found in Reference (Zhijian Liu, Li, Liu, et al., 2017). It can be clearly seen that though a large number of new designs are generated, they will be screened with the target criteria. Only a relatively small number of designs with the predicted results fulfill the criterions will be recorded in the database. Of course, sometimes it is hard to experimentally validate all these candidates after screening. It is recommended that picking at least two cases for experimental validation can ensure that the screening results are reliable.

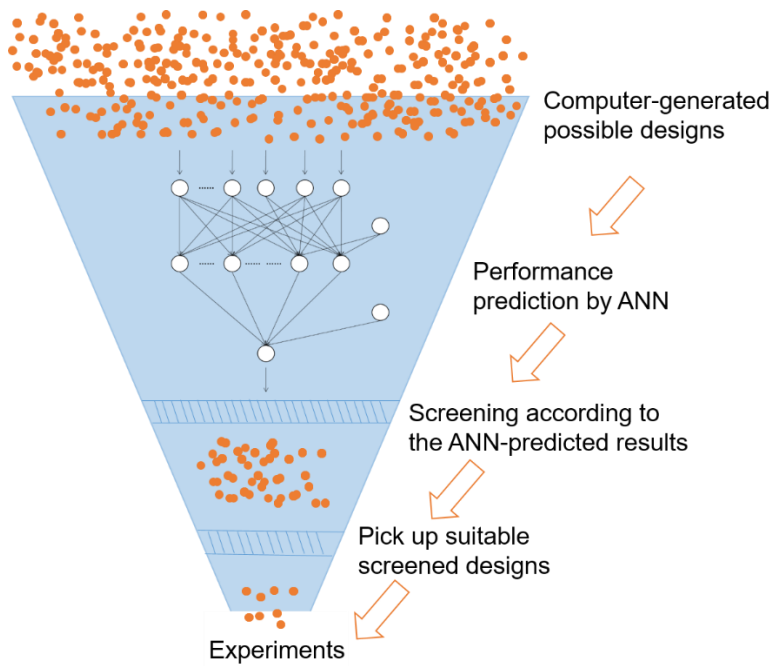

*Figure 9. An HTS process for solar energy system optimization. Each orange circle represents a possible design. The inset shows the schematic inter-connected structure of a general ANN algorithm. Reproduced with permission from Reference* (Li et al., 2017).

A key step of this HTS process is the generation of inputs for the new designs. Without a rational criterion for input generation, there will be infinite possible combinations of inputs, which will lead to infinite computations. For an ANN model, an interesting way is to generate the inputs according to the final weights of the network: the independent variables with larger weight contributions will be assigned more possible input values. The basic assumption here is simple: the independent variables with higher weight will lead to more significant changes to the dependent variables. It should be noted that the numerical weights do not contain any physical meaning, and the weights are usually different under different repeated training of a given ANN structure, since the initial weights of the ANN are usually selected randomly by the weight optimization algorithm. Fortunately, if each repeated ANN is well-trained, the weight values would be relatively stable in multiple trainings. This method provides a quick decision of input generations, which does not require people to know the exact physical meanings of the variables. However, sometimes people have to artificially assign more values to some independent variables. Taking the final temperature in this case as an example, it is not a part of the WGET-SWH installation, but it highly correlates with the heat collection rate. Thus, the temperature effect should not be ignored. More final temperature should be assigned than it is expected from its weight contributions. The number of each selected value for generation is shown in Table 5. All these values randomly combine with each other, constructing around $3.5 \times 10^8$ possible design combinations as the inputs of the ANN. Also, if people want to use other machine learning algorithms that do not require weight calculations (*e.g.*, SVM), the variables' physical meanings should be considered: the independent variables with more significant physical influences to the dependent variable should be assigned more selected values.

*Table 5. Number of selected values of different independent variables. Reproduced with permission from Reference* (Zhijian Liu, Li, Liu, et al., 2017).

|  | Tube Length (mm) | Number of Tubes | TCD (mm) | Tank Volume (kg) | Collector Area (m²) | Angle (°) | Final Temp. (°C) |
|---|---|---|---|---|---|---|---|
| Number of Selected Values | 5 | 30 | 5 | 111 | 50 | 5 | 17 |

Abbreviations: TCD: tube center distance, final temp.: final temperature. Tank volume was defined as the maximum mass of water in tank (kg).

## A General HTS Process for Energy System Optimization

So far, all the essential processes for the design and optimization of a high-performance WGET-SWH have been introduced, which mainly include two parts: i) developing a predictive model and ii) screening possible candidates. Here a general HTS framework (that might be used for other energy systems) will be introduced and discussed.

The proposed HTS framework for the design and optimization of energy system is shown in Figure 10. When all the preconditions of the "cylinders" shown in Figure 10 are fulfilled, a completed machine learning-assisted HTS process can be achieved. Since the final target of this HTS process is to discover new designed candidates with optimized performance, there should be a database that record all the independent and dependent variables of the new candidates for future use. It should be noted that these predicted candidates will have the independent variables different (or partially different) from the experimental database. Also, the results from the validation experiments should be added to the previous experimental database. By combining the validation experiment results with the previous measurement database, people can reconstruct a new experimental database for future applications. People can refer to either the predicted candidate database or the new experimental data for industrial or commercial use. This framework can be achieved by a machine learning code combined with some additional simple coding. Here, the authors expect that this framework not only works for optimizing WGET-SWH, but also works for the optimization cases of other energy systems.

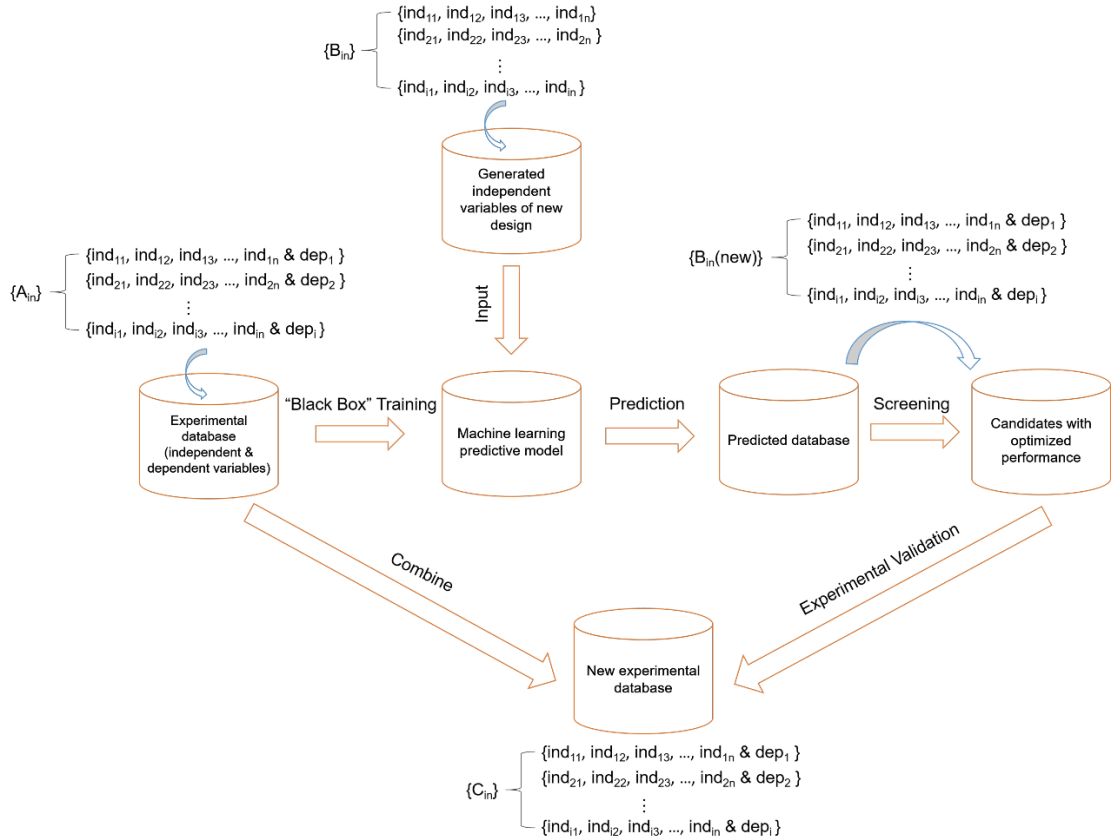

*Figure 10. A proposed framework of machine learning-assisted HTS process for target performance optimization. "int" is the independent variable. "dep" is the dependent variable. {$A_{in}$} is the original experimental database. {$B_{in}$} is the generated independent variables as the inputs. {$B_{in}(new)$} is the generated independent variables and their predicted dependent variables. {$C_{in}$} is the new experimental database combining the original experimental database and the experimental validation results of the screened candidates. Reproduced with permission from Reference* (Li et al., 2017).

## FUTURE RESEARCH DIRECTIONS

Though there is a great success on the prediction and optimization of SWHs using machine learning methods, there still remain some vital questions that should be addressed in the future: how to modify the HTS process for adopting different energy systems? Can people further simplify the HTS process and/or the generation rules of design inputs? Can people develop a user-friendly platform for the database of the designed candidates? Addressing these questions would be of great importance to make the HTS-based energy system optimization more applicable.

It should be noted that to achieve a successful HTS-assisted optimization of energy systems, both predictive model training and HTS process are necessary. Both of them require (more or less) some coding works, which meanwhile require some programming knowledge. If the thermal performance of an energy system can be easily optimized by the empirical knowledge, HTS process will no longer be recommended. Thus, the proposed HTS optimization process is only effective for the design of those energy systems with more complicated internal structures. This is the same as the basic mission of a machine learning model: to deal with those problems that are too complicated to be addressed by conventional methods. If things are simple, machine learning will no longer be cost-effective.

In the near future, it is expected to see that more commercial energy systems are optimized by a machine learning-based HTS process. With higher designed performances, it is expected that higher economic and environmental benefits can be achieved.

## CONCLUSION

In this Chapter, the authors have shown that machine learning techniques are powerful tools for predicting and optimizing the thermal performance of SWHs. Picking WGET-SWH as a case study, the authors have developed a knowledge-based measurement of thermal performances using the simple inputs measured by the "portable test instruments". Various machine learning models (ANN, SVM and ELM) were subsequently compared for the prediction of the heat collection rate and heat loss coefficient of WGET-SWH. To provide a more user-friendly measurement for practical use, an ANN-based software was developed in both PC and Android platforms.

To optimize the heat collection rate of a WGET-SWH, the authors have developed an ANN-based HTS process to screen around $3.5 \times 10^8$ possible design combinations. Candidates with high predicted heat collection rates were screened and recorded into a database for future use. Validation experiments on two selected cases in the candidate database showed surprisingly high thermal performances. All these results show that machine learning not only provides a strong predictive power for thermal performance prediction, but also provide a brand new insight for the performance optimization of an energy system. The authors also expect that this new HTS-based optimization strategy can be more widely used in the near future in the area of energy engineering.

## ACKNOWLEDGMENT

We are grateful to all the colleagues who involved in the data collection, programming and research discussions. This work was supported by the Major Basic Research Development and Transformation Program of Qinghai province (no. 2016-NN-141) and Natural Science Foundation of Hebei (no. E2017502051).